\documentclass[runningheads]{llncs}
\usepackage{graphicx}

\usepackage[T1]{fontenc}
\usepackage[outline]{contour}
\usepackage{xcolor}

\usepackage{times}
\usepackage{soul}
\usepackage{url}
\usepackage[utf8]{inputenc}
\usepackage{graphicx}
\usepackage{amsmath}
\usepackage{booktabs}
\usepackage{arydshln}
\usepackage{multirow}
\usepackage{makecell}
\usepackage{listings}
\usepackage{threeparttable}
\usepackage{algorithm} 
\usepackage{algorithmicx} 
\usepackage{amssymb}
\usepackage[noend]{algpseudocode}
\usepackage{footmisc}
\usepackage{hyperref}
\usepackage[misc]{ifsym}

\begin{document}
\title{Biomedical Knowledge Graph Embeddings \break with Negative Statements}
%
%
\author{Rita T. Sousa\inst{1}\orcidID{0000-0002-7241-8970} \and
Sara Silva\inst{1}\orcidID{0000-0001-8223-4799} \and
Heiko Paulheim\inst{2}\orcidID{0000-0003-4386-8195} \and 
Catia Pesquita\inst{1}\orcidID{0000-0002-1847-9393}}
\authorrunning{Sousa et al.}
\institute{LASIGE, Faculdade de Ciências da Universidade de Lisboa, Portugal \\
\email{\{risousa,sgsilva,clpesquita\}@ciencias.ulisboa.pt} \\
\and
Data and Web Science Group, Universität Mannheim, Germany \\ 
\email{heiko.paulheim@uni-mannheim.de}\\
}
\maketitle              
\begin{abstract}
A knowledge graph is a powerful representation of real-world entities and their relations. The vast majority of these relations are defined as positive statements, but the importance of negative statements is increasingly recognized, especially under an Open World Assumption. 
Explicitly considering negative statements has been shown to improve performance on tasks such as entity summarization and question answering or domain-specific tasks such as protein function prediction. However, no attention has been given to the exploration of negative statements by knowledge graph embedding approaches despite the potential of negative statements to produce more accurate representations of entities in a knowledge graph.

We propose a novel approach, TrueWalks, to incorporate negative statements into the knowledge graph representation learning process. 
In particular, we present a novel walk-generation method that is able to not only differentiate between positive and negative statements but also take into account the semantic implications of negation in ontology-rich knowledge graphs. This is of particular importance for applications in the biomedical domain, where the inadequacy of embedding approaches regarding negative statements at the ontology level has been identified as a crucial limitation.

We evaluate TrueWalks in ontology-rich biomedical knowledge graphs in two different predictive tasks based on KG embeddings: protein-protein interaction prediction and gene-disease association prediction. We conduct an extensive analysis over established benchmarks and demonstrate that our method is able to improve the performance of knowledge graph embeddings on all tasks. 

\end{abstract}

\section{Introduction}

Knowledge Graphs (KGs) represent facts about real-world entities and their relations and have been extensively used to support a range of applications from question-answering and recommendation systems to machine learning and analytics~\cite{hogan2021knowledge}. 
KGs have taken to the forefront of biomedical data through their ability to describe and interlink information about biomedical entities such as genes, proteins, diseases and patients, structured according to biomedical ontologies. This supports the analysis and interpretation of biological data, for instance, through the use of semantic similarity measures~\cite{pesquita2009semantic}. More recently, a spate of KG embedding methods~\cite{wang2017knowledge} have emerged in this space and have been successfully employed in a number of biomedical applications~\cite{mohamed2021biological}. 
The impact of KG embeddings in biomedical analytics is expected to increase in tandem with the growing volume and complexity of biomedical data. However, this success relies on the expectation that KG embeddings are semantically meaningful representations of the underlying biomedical entities.  

Regardless of their domain, the vast majority of KG facts are represented as positive statements, e.g. \textit{$(\text{hemoglobin}, \text{hasFunction}, \text{oxygen transport})$}. Under a Closed World Assumption, negative statements are not required, since any missing fact can be assumed as a negative. However, real-world KGs reside under the Open World Assumption where non-stated negative facts are formally indistinguishable from missing or unknown facts, which can have important implications across a variety of tasks. 

The importance of negative statements is increasingly recognized~\cite{arnaout_negative_2021-1,flouris2006inconsistencies}.
For example, in the biomedical domain, the knowledge that a patient does not exhibit a given symptom or a protein does not perform a specific function is crucial for both clinical decision-making and biomedical insight.
While ontologies are able to express negation and the enrichment of KGs with interesting negative statements is gaining traction, existing KG embedding methods are not able to adequately utilize them~\cite{kulmanov2021semantic}, which ultimately results in less accurate representations of entities.

\begin{figure}[t!]
    \centering
    \includegraphics[width=0.9\textwidth]{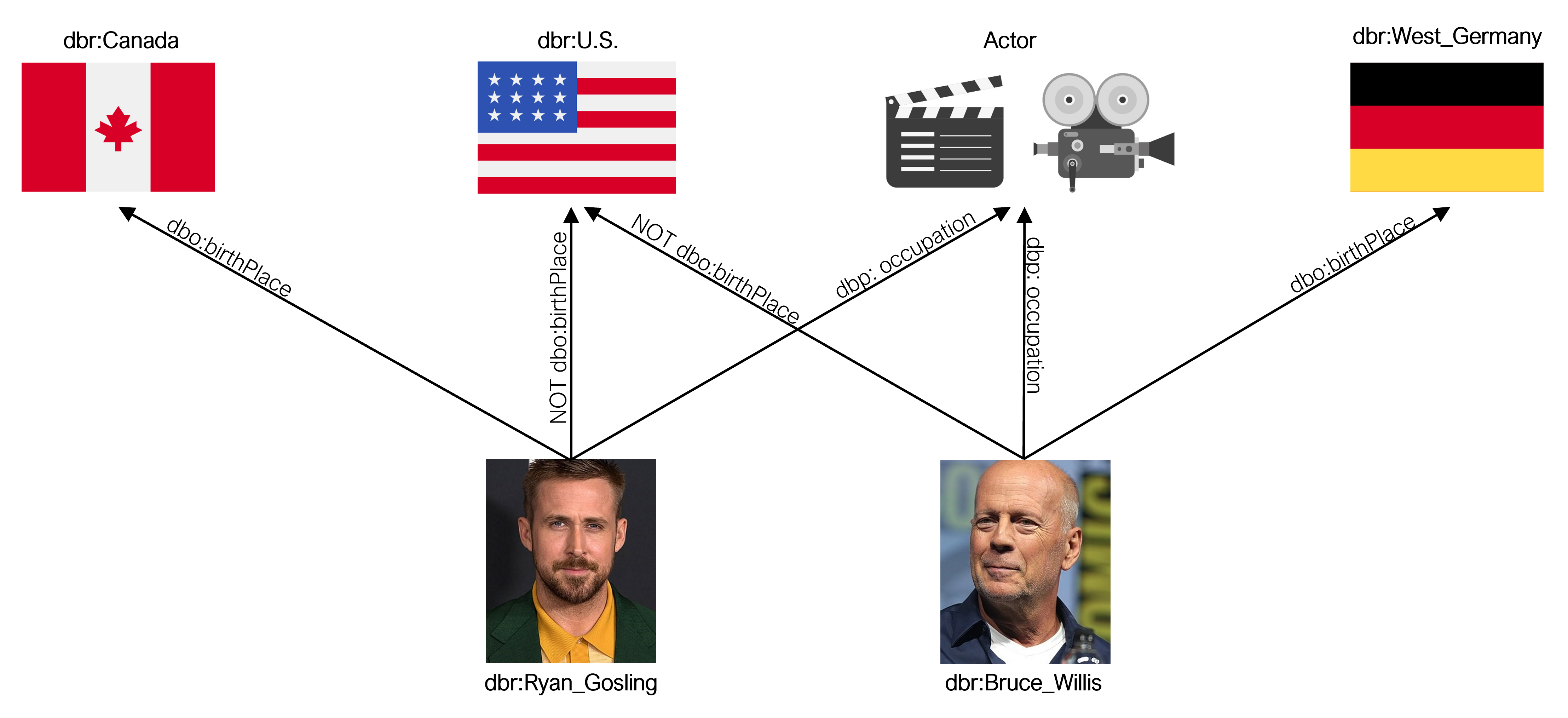}
    \caption{A DBPedia example motivating the negative statements problem. The author of Bruce Willis' picture is Gage Skidmore.}
    \label{fig:dbpediaexample}
\end{figure}

We propose True Walks, to the best of our knowledge, the first-ever approach that is able to incorporate negative statements into the KG embedding learning process. 
This is fundamentally different from other KG embedding methods, which produce negative statements by negative random sampling strategies to train representations that bring the representations of nodes that are linked closer, while distancing them from the negative examples. TrueWalks uses explicit negative statements to produce entity representations that take into account both existing attributes and lacking attributes.
For example, for the negative statement \textit{$(\text{Bruce Willis}, \text{NOT birthPlace}, \text{U.S.})$}, our representation would be able to capture the similarity between Bruce Willis and Ryan Gosling, since neither was born in the U.S (see Figure~\ref{fig:dbpediaexample}). The explicit declaration of negative statements such as these is an important aspect of more accurate representations, especially when they capture unexpected negative statements (i.e., most people would expect that both actors are U.S. born). Using TrueWalks, Bruce Willis and Ryan Gosling would be similar not just because they are both actors but also because neither was born in the U.S.

True Walks generates walks that can distinguish between positive and negative statements and consider the semantic implications of negation in KGs that are rich in ontological information, particularly in regard to inheritance.
This is of particular importance for applications in the biomedical domain, where the inadequacy of embedding approaches regarding negative statements has been identified as a crucial limitation~\cite{kulmanov2021semantic}. We demonstrate that the resulting embeddings can be employed to determine semantic similarity or as features for relation prediction. We evaluate the effectiveness of our approach in two different tasks, protein-protein interaction prediction and gene-disease association prediction, and show that our method improves performance over state-of-the-art embedding methods and popular semantic similarity measures.

Our contributions are as follows:
\begin{itemize}
\item We propose TrueWalks, a novel method to generate random walks on KGs that are aware of negative statements and results in the first KG embedding approach that considers negative statements.
\item We develop extensions of popular path-based KG embedding methods implementing the TrueWalks approach.
\item We enrich existing KGs with negative statements and propose benchmark datasets for two popular biomedical KG applications: protein-protein interaction (PPI) prediction and gene-disease association (GDA) prediction.
\item We report experimental results that demonstrate the superior performance of TrueWalks when compared to state-of-the-art KG embedding methods.
\end{itemize}

\section{Related Work}

\subsection{Exploring Negative Statements}

Approaches to enrich existing KGs with interesting negative statements have been proposed both for general-purpose KGs such as Wikidata~\cite{arnaout_wikinegata_2021} and for domain-specific ones such as the Gene Ontology (GO)~\cite{fu_neggoa_2016,vesztrocy2020benchmarking}.
Exploring negative statements has been demonstrated to improve the performance of various applications. \cite{arnaout_negative_2021-1} developed a method to enrich Wikidata with interesting negative statements and its usage improved the performance on entity summarization and decision-making tasks. \cite{vesztrocy2020benchmarking} have designed a method to enrich the GO~\cite{GO2018} with relevant negative statements indicating that a protein does not perform a given function and demonstrated that a balance between positive and negative annotations supports a more reasonable evaluation of protein function prediction methods. Similarly, \cite{fu_neggoa_2016} enriched the GO with negative statements and demonstrated an associated increase in protein function prediction performance. The relevance of negative annotations has also been recognized in the prediction of gene-phenotype associations in the context of the Human-Phenotype Ontology (HP)~\cite{HP2021}, but the topic remains unexplored \cite{liu2021computational}.
It should be highlighted that KG embedding methods have not been employed in any of these approaches to explore negative statements.

\subsection{Knowledge Graph Embeddings}

KG embedding methods map entities and their relations expressed in a KG into a lower-dimensional space while preserving the underlying structure of the KG and other semantic information~\cite{wang2017knowledge}.
These entity and relation embedding vectors can then be applied to various KG applications such as link prediction, entity typing, or triple classification. 
In the biomedical domain, KG embeddings have been used in machine learning-based applications in which they are used as input in classification tasks or to predict relations between biomedical entities.
\cite{kulmanov2021semantic} provides an overview of KG embedding-based approaches for biomedical applications.

Translational models, which rely on distance-based scoring functions, are some of the most widely employed KG embedding methods. 
A popular method, TransE~\cite{bordes2013translating}, assumes that if a relation holds between two entities, the vector of the head entity plus the relation vector should be close to the vector of the tail entity in the vector space.
TransE has the disadvantage of not handling one-to-many and many-to-many relationships well. To address this issue, TransH~\cite{wang2014knowledge} introduces a relation-specific hyperplane for each relation and projects the head and tail entities into the hyperplane. 
TransR~\cite{lin2015learning} builds entity and relation embeddings in separate entity space and relation spaces. 

Semantic matching approaches are also well-known and use similarity-based scoring functions to capture the latent semantics of entities and relations in their vector space representations.
For instance, DistMult~\cite{yang2014embedding} employs tensor factorization to embed entities as vectors and relations as diagonal matrices.

\subsection{Walk-Based Embeddings}

More recently, random walk-based KG embedding approaches have emerged. These approaches are built upon two main steps: (i) producing entity sequences from walks in the graph to produce a corpus of sequences that is akin to a corpus of word sequences or sentences; (2) using those sequences as input to a neural language model~\cite{mikolov2013efficient} that learns a latent low-dimensional representation of each entity within the corpus of sequences. 

DeepWalk~\cite{perozzi2014deepwalk} first samples a set of paths from the input graph using uniform random walks. Then it uses those paths to train a skip-gram model, originally proposed by the word2vec approach for word embeddings~\cite{mikolov2013efficient}.
Node2vec~\cite{grover2016node2vec} introduces a different biased strategy for generating random walks and exploring diverse neighborhoods. The biased random walk strategy is controlled by two parameters: the likelihood of visiting immediate neighbors (breadth-first search behavior), and the likelihood of visiting entities that are at increasing distances (depth-first search behavior). 
Neither DeepWalk nor node2vec take into account the direction or type of the edges. Metapath2vec~\cite{dong2017metapath2vec} proposes random walks driven by metapaths that define the node type order by which the random walker explores the graph.
RDF2Vec~\cite{ristoski2016rdf2vec} is inspired by the node2vec strategy but it considers both edge direction and type making it particularly suited to KGs. OWL2Vec*~\cite{chen2021owl2vec} was designed to learn ontology embeddings and it also employs direct walks on the graph to learn graph structure.

\subsection{Tailoring Knowledge Graph Embeddings}

Recent KG embedding approaches aim to tailor representations by considering different semantic, structural or lexical aspects of a KG and its underlying ontology. Approaches such as EL~\cite{kulmanov2019embeddings} and BoxEL~\cite{xiong2022faithful} embeddings are geometric approaches that account for the logical structure of the ontology (e.g., intersection, conjunction, existential quantifiers). OWL2Vec*~\cite{chen2021owl2vec} and OPA2Vec~\cite{smaili2019opa2vec} take into consideration the lexical portion of the KG (i.e., labels of entities) when generating graph walks or triples. OPA2Vec also offers the option of using a pre-trained language model to bootstrap the KG embedding.  
Closer to our approach, OLW2Vec* contemplates the declaration of inverse axioms to enable reverse path traversal, however, this option was found lacking for the biomedical ontology GO. Finally, different approaches have been proposed to train embeddings that are aware of the order of entities in a path, such as ~\cite{zhu2019representation} and ~\cite{portisch2021putting}, which extend TransE and RDF2Vec, respectively.

\section{Methods}

\subsection{Problem Formulation}

\begin{figure}[b!]
    \centering
    \includegraphics[width=0.55\textwidth]{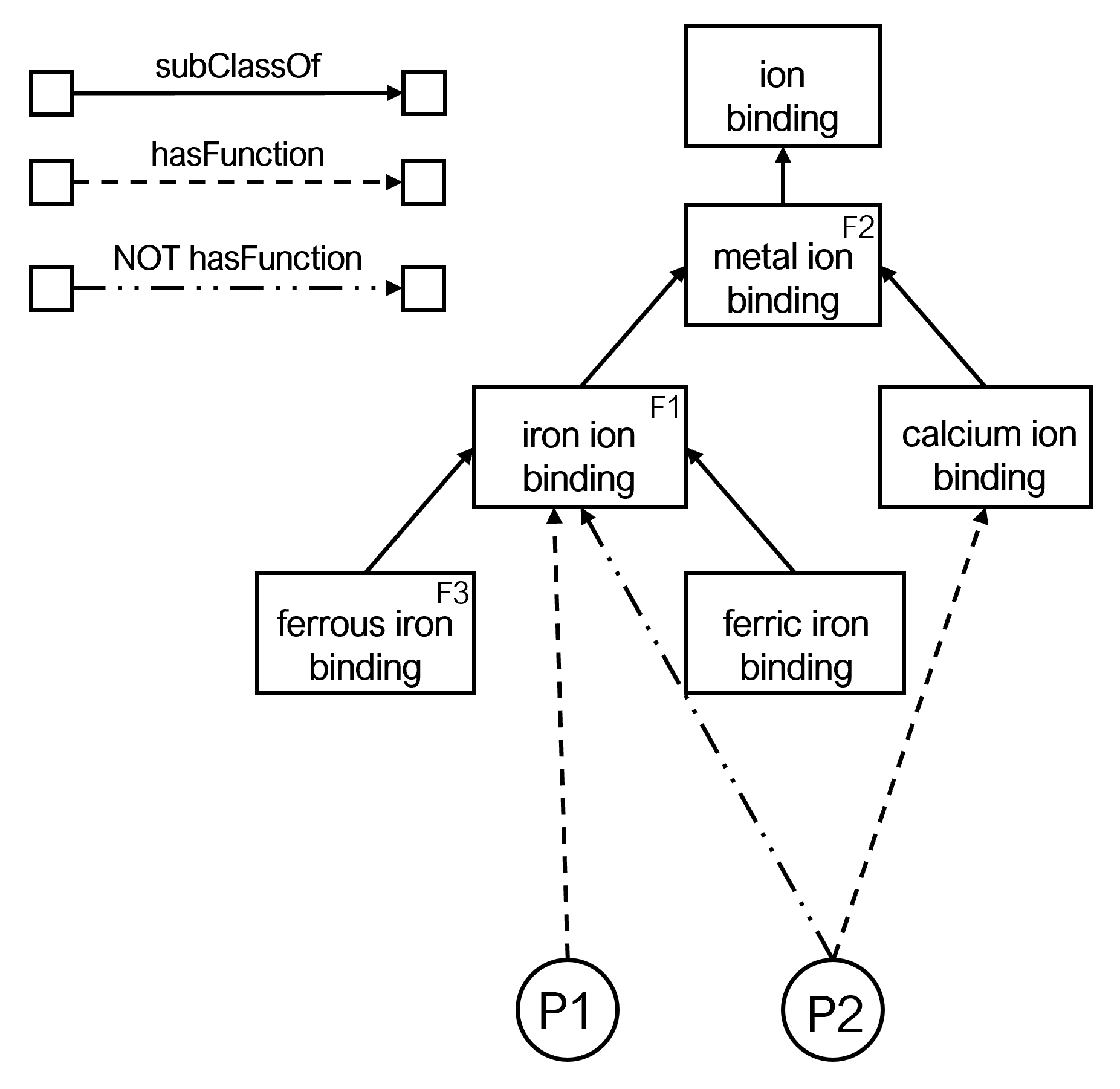}
    \caption{A GO KG subgraph motivating the \textit{reverse inheritance} problem.}
    \label{fig:example}
\end{figure}

In this work, we address the task of learning a relation between two KG entities (which can belong to the same or different KGs) when the relation itself is not encoded in the KG. We employ two distinct approaches: 
(1) using the KG embeddings of each entity as features for a machine learning algorithm and (2) comparing the KG embeddings directly through a similarity metric. 

We target ontology-rich KGs that use an ontology to provide rich descriptions of real-world entities instead of focusing on describing relations between entities themselves. These KGs are common in the biomedical domain. As a result, the KG's richness lies in the TBox, with a comparatively less complex ABox, since entities have no links between them.
We focus on Web Ontology Language (OWL)~\cite{grau2008owl} ontologies since biomedical ontologies are typically developed in OWL or have an OWL version.

Biomedical entities in a KG are typically described through positive statements that link them to an ontology. For instance, to state that a protein $P$ performs a function $F$ described under the GO, a KG can declare the axiom $P \sqsubseteq \exists hasFunction.F$. However, the knowledge that a given protein does not perform a function can also be relevant, especially to declare that a given protein does not have an activity typical of its homologs~\cite{gaudet2017gene}. Likewise, the knowledge that a given disease does not exhibit a particular phenotype is also decisive in understanding the relations between diseases and genes~\cite{liu2021computational}.
We consider the definition of grounded negative statements proposed by~\cite{arnaout_negative_2021-1} as $ \neg (s, p, o)$ which is satisfied if $(s, p, o) \notin KG$ and expressed as a \mbox{\textit{NegativeObjectPropertyAssertion}}\footnote{https://www.w3.org/TR/owl2-syntax/\#Negative\_Object\_Property\_Assertions}.
Similar to what was done in~\cite{arnaout_negative_2021-1}, we do not have a negative object property assertion for every missing triple. Negative statements are only included if there is clear evidence that a triple does not exist in the domain being captured. Taking the protein example, negative object property assertions only exist when it has been demonstrated that a protein does not perform a particular function.

An essential difference between a positive and a negative statement of this kind is related to the implied inheritance of properties exhibited by the superclasses or subclasses of the assigned class. 
Let us consider that \textit{$(P_1, \text{hasFunction}, F_1)$} and \textit{$(F_1, \text{subClassOf},F_2)$}. This implies that  \textit{$(P_1, \text{hasFunction}, F_2)$}, since an individual with a class assignment also belongs to all superclasses of the given class, e.g., a protein that performs \textit{iron ion binding} also performs \textit{metal ion binding} (see Figure~\ref{fig:example}). This implication is easily captured by directed walk generation methods that explore the declared subclass axioms in an OWL ontology. 
However, when we have a negative statement, such as \textit{$\neg (P_2, \text{hasFunction}, F_1)$}, it does not imply that \textit{$\neg (P_2, \text{hasFunction}, F_2)$}. There are no guarantees that a protein that does not perform \textit{iron ion binding} also does not perform \textit{metal ion binding}, since it can very well, for instance, perform \textit{calcium ion binding}.
However, for \textit{$(F_3, \text{subClassOf}, F_1)$} the negative statement \textit{$ \neg (P_2, \text{hasFunction}, F_1)$} implies that \textit{$\neg (P_2, \text{hasFunction}, F_3)$}, as a protein that does not perform \textit{iron ion binding} also does not perform \textit{ferric iron binding} nor \textit{ferrous iron binding}. 
Therefore, we need to be able to declare that protein $P_1$ performs both functions $F_1$ and $F_3$, but that  $P_2$ performs $F_1$ but not $F_3$. 
Since OWL ontologies typically declare subclass axioms, there is no opportunity for typical KG embedding methods to explore the reverse paths that would more accurately represent a negative statement.

The problem we tackle is then two-fold: how can the \textit{reverse inheritance} implied by negative statements be adequately explored by walk-based KG embedding methods, and how can these methods distinguish between negative and positive statements.

\subsection{Overview}

\begin{figure}[!t]
    \centering
    \includegraphics[width=0.65\textwidth]{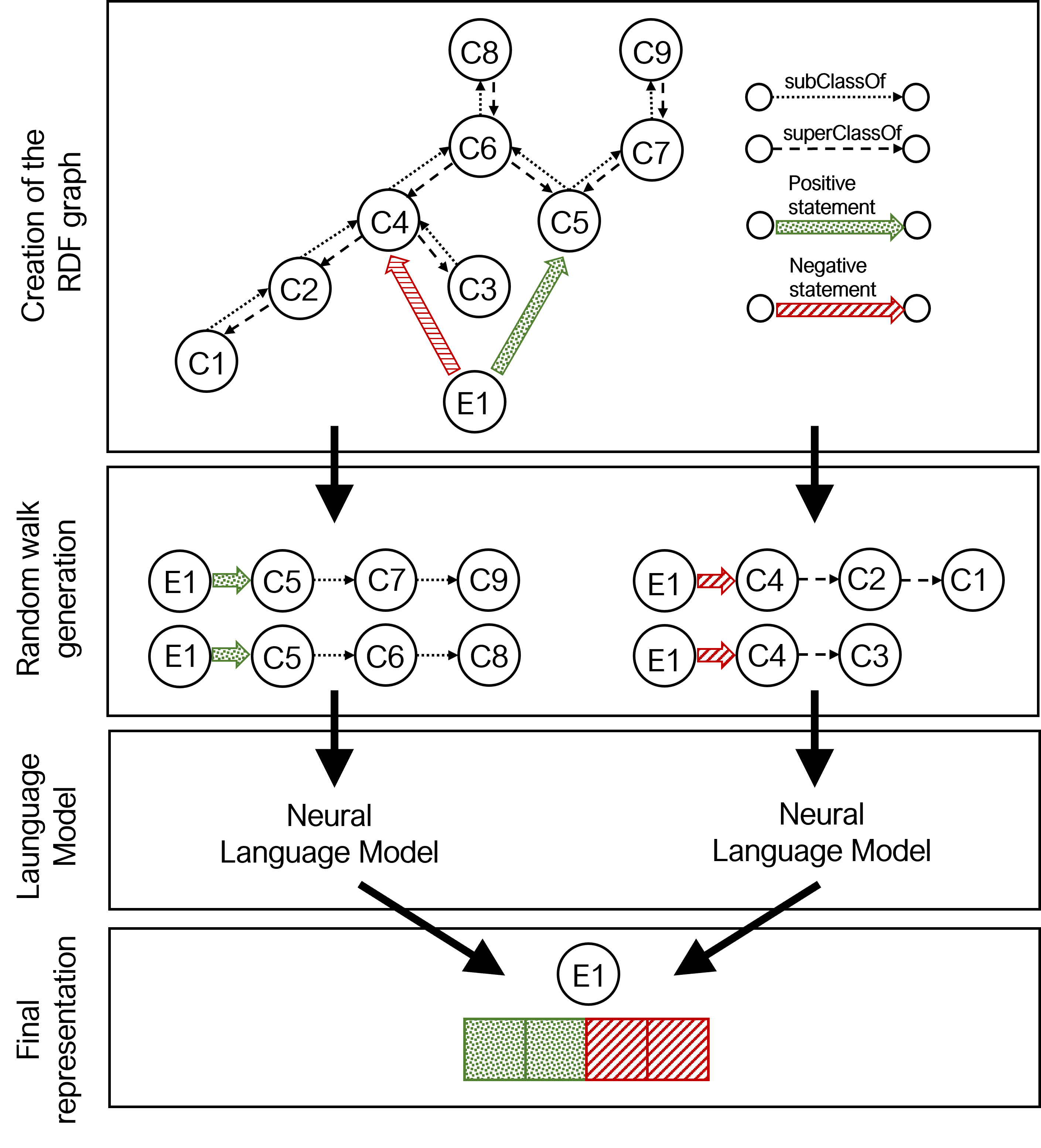}
    \caption{Overview of the TrueWalks method with the four main steps: (i) creation of the RDF graph, (ii) random walk generation with negative statements; (iii) neural language models, and (iv) final representation.}
    \label{fig:TrueWalks}
\end{figure}

An overview of TrueWalks, the method we propose, is shown in Figure~\ref{fig:TrueWalks}. The first step is the transformation of the KG into an RDF Graph. Next, our novel random walk generation strategy that is aware of positive and negative statements is applied to the graph to produce a set of entity sequences. The positive and negative entity walks are fed to neural language models to learn a dual latent representation of the entities. TrueWalks has two variants: one that employs the classical skip-gram model to learn the embeddings (TrueWalks), and one that employs a variation of skip-gram that is aware of the order of entities in the walk (TrueWalksOA, i.e. order-aware). 

\subsection{Creation of the RDF Graph}

The first step is the conversion of an ontology-rich KG into an RDF graph. This is a directed, labeled graph, where the edges represent the named relations between two resources or entities, represented by the graph nodes\footnote{https://www.w3.org/RDF/}. We perform the transformation according to the \textit{OWL to RDF Graph Mapping} guidelines defined by the W3C\footnote{https://www.w3.org/TR/owl2-mapping-to-rdf/}. Simple axioms can be directly transformed into RDF triples, such as subsumption axioms for atomic entities or data and annotation properties associated with an entity. Axioms involving complex class expressions are transformed into multiple triples which typically require blank nodes. 

Let us consider the following existential restriction of the class \textit{obo:GO\_0034708 (methyltransferase complex)} that encodes the fact that a methyltransferase complex is part of at least one intracellular anatomical structure:
\par\textit{ObjectSomeValuesFrom(obo:BFO\_0000050 (part of), \newline 
\indent obo:GO\_0005622 (intracellular anatomical structure))}
\par
\noindent Its conversion to RDF results in three triples: \par
\textit{(obo:GO\_0034708, rdfs:subClassOf, \_:x)}
\par
 \textit{(\_:x, owl:someValuesFrom,obo:GO\_0005622)}
 \par
\textit{(\_:x, owl:onProperty,obo:BFO\_0000050)}
\par
\noindent where \_:x denotes a blank node.

\subsection{Random Walk Generation with Negative Statements} \label{randomwalk}

The next step is to generate the graph walks that will make up the corpus (see Algorithm \ref{alg:TW}). For a given graph $G = (V, E)$  where $E$ is the set of edges and $V$ is the set of vertices, for each vertex $v_r \in V_r$, where $V_r$ is the subset of individuals for which we want to learn representations, we generate up to $w$ graph walks of maximum depth $d$ rooted in vertex $v_r$. 
We employ a depth-first search algorithm, extending on the basic approach in \cite{ristoski2016rdf2vec}. At the first iteration, we can find either a positive or negative statement. From then on, walks are biased: a positive statement implies that whenever a subclass edge is found it is traversed from subclass to superclass, whereas a negative statement results in a traversal of subclass edges in the opposite direction (see also Figure~\ref{fig:TrueWalks}). This generates paths that follow the pattern $v_r \rightarrow e_{1i} \rightarrow v_{1i} \rightarrow e_{2i}$.
The set of walks is split in two, negative statement walks and positive statement walks. This will allow the learning of separate latent representations, one that captures the positive aspect and one that captures the negative aspect. 

\begin{algorithm}[!t]
\algrenewcommand\algorithmicindent{0.5em}%
\caption{Walk generation for one entity using TrueWalks. The function GET NON VISITED NEIGHBOURS(status) is used to generate the random walks using a depth-first search. It gets the neighbors of a given node that have not yet been visited in previous iterations. If the status is negative (which means that the first step in the walk was made with a negative statement), the neighbors will include all the non-visited neighbors except those connected through subclass statements, and if the status is positive, it will include all the neighbors except those connected through superclass statements.}
 \begin{algorithmic}[1]
 \State $d \gets max\_depth\_walks$
  \State $w \gets max\_number\_of\_walks$
  \State $ent \gets root\_entity$
 \Function{get TrueWalks}{$ent$}
 \State $pos\_walks \gets \Call{get random walks}{ent,positive}$
  \State $neg\_walks \gets \Call{get random walks}{ent,negative}$
 \State \Return $pos\_walks, neg\_walks$
 \EndFunction

\Function{get random walks}{$ent, status$}
  \While{$len(walks) < w$} 
  \State $walk \gets ent$
   \State{$depth \gets 1$}
    \While{ $depth < d$} 
    \State $last \gets len(walk)==d$ 
    \State $e,v \gets \Call{get neighbor}{walk,status,last}$
    \If{$e,v == None$}
    \State break
    \EndIf
     \State $walk$.append($e,v$) 
    \State $depth++$
    \EndWhile
 \State $walks$.append($walk$)
  \EndWhile
 \State \Return $walks$
\EndFunction

 \Function{get neighbor}{$walk,status,last$}
  \State $n \gets \Call{get non visited neighbors}{status} $
  \If{$len(n) == 0 \And len(walk) >2$}
  \State $e,v \gets  walk[-2], walk[-1]$
 \State \Call{add visited neighbors}{$e,v,len(walk)-2,status$}
  \State \Return $None$
  \EndIf
  \State $e,v \gets n[rand()]$
  \If{$last$}
  \State \Call{add visited neighbors}{$e,v,len(walk),status$}
   \EndIf
   \State \Return $e,v$
   \EndFunction

 \end{algorithmic}
 \label{alg:TW}
 \end{algorithm}

An important aspect of our approach is that, since OWL is converted into an RDF graph for walk-based KG embedding methods, a negative statement declared using a simple object property assertion (e.g. \textit{notHasFunction}) could result in the less accurate path: \textit{Protein P $\rightarrow$ notHasFunction $\rightarrow$ iron ion binding  $\rightarrow$ subClassOf $\rightarrow$ ion binding}. Moreover, random walks directly over the \textit{NegativeObjectPropertyAssertion}, since it is decomposed into multiple triples using blank nodes, would also result in inaccurate paths. However, our algorithm produces more accurate paths, e.g.: \textit{Protein P $\rightarrow$ notHasFunction $\rightarrow$ iron ion binding $\rightarrow$ superClassOf $\rightarrow$ ferric iron binding} by adequately processing the \mbox{\textit{NegativeObjectPropertyAssertion}}.

\subsection{Neural Language Models}

We employ two alternative approaches to learn a latent representation of the individuals in the KG. For the first approach, we use the skip-gram model~\cite{mikolov2013efficient}, which predicts the context (neighbor entities) based on a target word, or in our case a target entity. 

Let $f : E \rightarrow \mathbb{R}^{d}$ be the mapping function from entities to the latent representations we will be learning, where $d$ is the number of dimensions of the representation ($f$ is then a matrix $|E| \times d$). Given a context window $c$, and a sequence of entities $e_1, e_2, e_3, ..., e_L$, the objective of the skip-gram model is to maximize the average log probability $p$:

\begin{equation}
 \frac{1}{L} \sum_{l=1}^{L}  \log p(e_{l+c} | e_{l})
\end{equation}
\noindent where $p(e_{l+c} | e_{l})$ is calculated using the softmax function:

\begin{equation}
    p(e_{l+c} | e_{l}) = \frac{\exp(f(e_{l+c}) \cdot f({e_l}))}{\sum_{e=1}^{E} \exp(f({e}) \cdot f({e_l}))}
\end{equation}
\noindent where $f(e)$ is the vector of the entity $e$. 

To improve computation time, we employ a negative sampling approach based in~\cite{mikolov2013efficient} that minimizes the number of comparisons required to distinguish the target entity, by taking samples from a noise distribution using logistic regression, where there are $k$ negative samples for each entity.

The second approach is the structured skip-gram model~\cite{ling2015two}, a variation of skip-gram that is sensitive to the order of words, or in our case, entities in the graph walks. The critical distinction of this approach is that, instead of using a single matrix $f$, it creates $c \times 2$ matrices,  $f_{-c},...,f_{-2},f_{-1},f_1,...,f_c$, each dedicated to predicting a specific relative position to the entity. To make a prediction  $p(e_{l+c} | e_{l})$, the method selects the appropriate matrix $f_l$. 

The neural language models are applied separately to the positive and negative walks, producing two representations for each entity. 

\subsection{Final Representations}

The two representations of each entity need to be combined to produce a final representation. Different vector operations can, in principle, be employed, such as the Hadamard product or the L1-norm. However, especially since we will employ these vectors as inputs for machine learning methods, we would like to create a feature space that allows the distinction between the negative and positive representations, motivating us to use a simple concatenation of vectors.

\section{Experiments}

We evaluate our novel approach on two biomedical tasks: protein-protein interaction (PPI) prediction and gene-disease association (GDA) prediction\cite{sousa2023benchmark}. 
These two challenges have significant implications for understanding the underlying mechanisms of biological processes and disease states. 

Both tasks are modeled as relation prediction tasks. For PPI prediction, we employ TrueWalks embeddings both as features for a supervised learning algorithm and directly for similarity-based prediction. For GDA prediction, since embeddings for genes and diseases are learned over two different KGs, we focus only on supervised learning. We employ a Random Forest algorithm across all classification experiments with the same parameters (see the supplementary file for details).

\subsection{Data}

Our method takes as input an ontology file, instance annotation file and a list of instance pairs.
We construct the knowledge graph (KG) using the RDFlib package~\cite{rdflib}, which parses the ontology file in OWL format and processes the annotation file to add edges to the RDFlib graph. The annotation file contains both positive and negative statements which are used to create the edges in the graph.

\begin{table}[!t]
\centering
\setlength{\tabcolsep}{4pt}
\footnotesize
\caption{Statistics for each KG regarding classes, instances, nodes, edges, positive and negative statements.}
\begin{tabular}{lrrr}
\toprule
& GO\textsubscript{PPI} & GO\textsubscript{GDA} & HP\textsubscript{GDA}\\ \midrule
Classes & 50918 & 50918 & 17060 \\
Literals and blank nodes & 532373 & 532373 & 442246\\
Edges & 1425102 & 1425102 & 1082859\\
Instances & 440 & 755 &  162 \\
Positive statements& 7364 & 10631 & 4197 \\
Negative statements & 8579  & 8966 & 225\\
\bottomrule
\end{tabular}
\label{tab:statistics} 
\end{table}

\subsubsection{Protein-Protein Interaction Prediction}

Predicting protein-protein interactions is a fundamental task in molecular biology that can explore both sequence and functional information~\cite{hu2021survey}. Given the high cost of experimentally determining PPI, computational methods have been proposed as a solution to the problem of finding protein pairs that are likely to interact and thus provide a selection of good candidates for experimental analysis. In recent years, a number of approaches for PPI prediction based on functional information as described by the GO have been proposed~\cite{zhang2016protein,kulmanov2019embeddings,smaili2019opa2vec,sousa2020evolving,kulmanov2021semantic}. The GO contains over 50000 classes that describe proteins or genes according to the molecular functions they perform, the biological processes they are involved in, and the cellular components where they act. 

The GO KG is built by integrating three sources: the GO itself~\cite{GO2018}, the Gene Ontology Annotation (GOA) data~\cite{gene2021gene}, and negative GO annotations~\cite{vesztrocy2020benchmarking} (details on the KG building method and data sources are available in the supplementary file). 
A GO annotation associates a Uniprot protein identifier with a GO class that describes it.
We downloaded the GO annotations corresponding to positive statements from the 
GOA database for human species. For each protein $P$ in the PPI dataset and each of its association statements to a function $F$ in GOA, we add the assertion \textit{$(P, \textit{hasFunction}, F)$}. 
We employ the negative GO associations produced in~\cite{vesztrocy2020benchmarking}, which were derived from expert-curated annotations of protein families on phylogenetic trees. For each protein $P$ in the PPI dataset and each of its association statements to a function $F$ in the negative GO associations dataset, 
we add a negative object property assertion. To do so, we use metamodeling (more specifically, punning \footnote{https://www.w3.org/TR/owl2-new-features/\#F12:\_Punning}) and represent each ontology class as both a class and an individual. This situation translates into using the same IRI. Then, we use a negative object property assertion to state that the individual representing a biomedical entity is not connected by the object property expression to the individual representing an ontology class. Table \ref{tab:statistics} presents the GO KG statistics.

The target relations to predict are extracted from the STRING database~\cite{STRING2021}. We considered the following criteria to select protein pairs: (i) protein interactions must be extracted from curated databases or experimentally determined (as opposed to computationally determined); (ii) interactions must have a confidence score above 0.950 to retain only high confidence interaction; (iii) each protein must have at least one positive GO association and one negative GO association. The PPI dataset contains 440 proteins, 1024 interacting protein pairs, and another 1024 pairs generated by random negative sampling over the same set of proteins.

\subsubsection{Gene-Disease Association Prediction}

Predicting the relation between genes and diseases is essential to understand disease mechanisms and identify potential biomarkers or therapeutic targets~\cite{eilbeck2017settling}. However, validating these associations in the wet lab is expensive and time-consuming, which fostered the development of computational approaches to identify the most promising associations to be further validated.  Many of these explore biomedical ontologies and KGs~\cite{vanunu2010,zakeri2018,robinson2014,asif2018,Luo2019} and some recent approaches even apply KG embedding methods such as DeepWalk~\cite{alshahrani2017neuro} or OPA2Vec~\cite{smaili2019opa2vec,nunes2021predicting}.

For GDA prediction, we have used the GO KG, the Human Phenotype Ontology (HP) KG (created from the HP file and HP annotations files), and a GDA dataset.
Two different ontologies are used to describe each type of entity. Diseases are described under the HP and genes under the GO. 
We built GO KG in the same fashion as in the PPI experiment, but instead of having proteins linked to GO classes, we have genes associated with GO classes. 
Regarding HP KG, HP~\cite{HP2021} describes phenotypic abnormalities found in human hereditary diseases.
The HP annotations link a disease to a specific class in the HP through both positive and negative statements. 

The target relations to predict are extracted from \mbox{DisGeNET}~\cite{pinero2019disgenet}, adapting the approach described in~\cite{nunes2021predicting} to consider the following criterion: each gene (or disease) must have at least one positive GO (or HP) association and one negative GO (or HP) association. 
This resulted in 755 genes, 162 diseases, and 107 gene-disease relations. To create a balanced dataset, we sampled random negative examples over the same genes and diseases. Table~\ref{tab:statistics} describes the created KGs. 

\subsection{Results and Discussion}

We compare TrueWalks against ten state-of-the-art KG embedding methods: 
TransE, TransH, TransR, ComplEx, distMult, DeepWalk, node2vec, metapath2vec, OWL2Vec* and RDF2Vec. TransE, TransH and TransR are representative methods of translational models. ComplEx and distMult are semantic matching methods. They represent a bottom-line baseline with well-known KG embedding methods. DeepWalk and node2vec are undirected random walk-based methods, and OWL2Vec* and RDF2Vec are directed walk-based methods. These methods represent a closer approach to ours, providing a potentially stronger baseline. Each method is run with two different KGs, one with only positive statements and one with both positive and negative statements. 
In this second KG, we declare the negative statements as an object property, so positive and negative statements appear as two distinct relation types.
The size of all the embeddings is 200 dimensions across all experiments (details on parameters can be found in the supplementary file), with TrueWalks generating two 100-dimensional vectors, one for the positive statement-based representation and one for the negative, which are concatenated to produce the final 200-dimensional representation. 

\begin{table*}[!b]
\footnotesize
\centering
\caption{Median precision, recall, and F-measure (weighted average F-measure) for PPI and GDA prediction. TrueWalks performance values are italicized/underlined when improvements are statistically significant with $p$-value $< 0.05$ for the Wilcoxon test against the positive (Pos)/positive and negative (Pos+Neg) variants of other methods. The best results are in bold.}
\begin{tabular}{clccccccc}
\toprule
& \multirow{2}{*}{\textbf{Method}} & \multicolumn{3}{c}{\textbf{PPI Prediction}} && \multicolumn{3}{c}{\textbf{GDA Prediction}}\\
\cmidrule{3-5} \cmidrule{7-9}
& & \textbf{Precision}   & \textbf{Recall} & \textbf{F-measure} & & \textbf{Precision}   & \textbf{Recall} & \textbf{F-measure} \\ \midrule
\parbox[t]{4mm}{\multirow{9}{*}{\rotatebox[origin=c]{90}{Pos}}} & TransE      & 0.553                & 0.546                & 0.554   &  & 0.533                & 0.538          & 0.531           \\
& TransH     & 0.566                & 0.562                & 0.566    &  & 0.556                & 0.563          & 0.548          \\
& TransR  & 0.620	& 0.607 & 0.616 && 0.594	& 0.600	& 0.592\\
& ComplEx & 0.680 & 0.659 & 0.679 && 0.597 & 0.625 & 0.598 \\
& distMult   & 0.765                & 0.737                & 0.754    &  & 0.585                & 0.600            & 0.575          \\
& DeepWalk & 0.813 & 0.836 & 0.822 &  & 0.618 & 0.646 & 0.629\\
& node2vec & 0.826 & 0.741  & 0.794 &  & 0.643 & 0.616 & 0.644\\
& metapath2vec & 0.562 & 0.563 & 0.561 & &  0.554 & 0.531 & 0.549\\
& OWL2Vec*   & 0.833                & 0.806                & 0.823    &  & 0.652                & 0.656 & 0.646          \\
& RDF2Vec    & 0.831                & 0.826                & 0.828    &  & 0.623                & 0.625          & 0.615          \\
 \midrule
\parbox[t]{4mm}{\multirow{11}{*}{\rotatebox[origin=c]{90}{Pos+Neg}}} & TransE       & 0.584                & 0.582           & 0.585  &   & 0.597                & 0.585          & 0.586           \\
& TransH      & 0.573                & 0.572           & 0.570  &   & 0.563                & 0.554          & 0.554           \\
& TransR & 0.722 & 0.678 & 0.704 && 0.633	& 0.625	& 0.630 \\
& ComplEx & 0.750 & 0.720 & 0.740 && 0.549 & 0.545 & 0.545\\
& distMult    & 0.813                & 0.740           & 0.784  &   & 0.530                & 0.523          & 0.534           \\
& DeepWalk &  0.843 & 0.834 &  0.841 & & 0.615 & 0.646 & 0.630 \\
& node2vec & 0.847 & 0.734 & 0.798 & &  0.614 & 0.594 & 0.621\\
& metapath2vec & 0.557 & 0.569 & 0.558  && 0.527 & 0.531 & 0.522\\
& OWL2Vec*    & 0.860                & 0.812           & 0.840  &  & 0.654                & 0.600          & 0.645            \\
& RDF2Vec     & 0.847                & \textbf{0.844}           & 0.845  &  & 0.625                & \textbf{0.661} & 0.630            \\
 \midrule
& TrueWalks   & \underline{\textbf{\textit{0.870}}} & 0.817           & \textit{0.846} &   & \underline{\textbf{\textit{0.667}}} & 0.625          & \underline{\textbf{\textit{0.661}}}            \\
& TrueWalksOA & \underline{\textit{0.868}}          & 0.836  & \underline{\textbf{\textit{0.858}}} & & \underline{\textit{0.661}}          & 0.616          & \underline{\textit{0.654}}\\ 
\bottomrule
\end{tabular}
\label{tab:Prediction}
\end{table*}

\subsubsection{Relation Prediction using Machine Learning}

To predict the relation between a pair of entities $e_1$ and $e_2$ using machine learning, we take their vector representations and combine them using the binary Hadamard operator to represent the pair: $ r(e_1, e_2) = v_{e_1} \times v_{e_2}$. 
The pair representations are then fed into a Random Forest algorithm for training using Monte Carlo cross-validation (MCCV)~\cite{XU20011}. MCCV is a variation of traditional $k$-fold cross-validation in which the process of dividing the data into training and testing sets (with $\beta$ being the proportion of the dataset to include in the test split) is repeated $M$ times. 
Our experiments use MCCV with $M=30$ and $\beta=0.3$. For each run, the predictive performance is evaluated based on recall, precision and weighted average F-measure. 
Statistically significant differences between TrueWalks and the other methods are determined using the non-parametric Wilcoxon test at \mbox{$p < $ 0.05}.

Table \ref{tab:Prediction} reports the median scores for both PPI and GDA prediction. The top half contains the results of the first experiment where we compare state-of-the-art methods using only the positive statements to TrueWalks (at the bottom) which uses both types. The results reveal that the performance of TrueWalks is significantly better than the other methods, improving both precision and F-measure. An improvement in precision, which is not always accompanied by an increase in recall, confirms the hypothesis that embeddings that consider negative statements produce more accurate representations of entities, which allows a better distinction of true positives from false positives.

A second experiment employs a KG with both negative and positive statements for all methods. 
Our method can accurately distinguish between positive statements and negative statements, as discussed in subsection~\ref{randomwalk}. For the remaining embedding methods, we declare the negative statements as an object property so that these methods distinguish positive and negative statements as two distinct types of relation.
This experiment allows us to test whether TrueWalks, which takes into account the positive or negative status of a statement, can improve the performance of methods that handle all statements equally regardless of status. 

The bottom half of Table~\ref{tab:Prediction} shows that both variants of TrueWalks improve on precision and F-measure for both tasks when compared with the state-of-the-art methods using both positive and negative statements.
This experiment further shows that the added information given by negative statements generally improves the performance of most KG embedding methods. However, no method surpasses TrueWalks, likely due to its ability to consider the semantic implications of inheritance and walk direction, especially when combined with the order-aware model. 

Comparing the two variants of TrueWalks demonstrates that order awareness does not improve performance in most cases. However, TrueWalksOA improves on precision and F-measure for all other state-of-the-art methods. These results are not unexpected since the same effect was observed in other order-aware embedding methods~\cite{portisch2021putting}.

Regarding the statistical tests, TrueWalks performance values are italicized/underlined in Table~\ref{tab:Prediction} when improvements over all other methods are statistically significant, except when comparing TrueWalks with OWL2Vec* for GDA, since in this particular case the improvement is not statistically significant.

\subsubsection{Relation Prediction using Semantic Similarity}

We also evaluate all methods in PPI prediction using KG embedding-based semantic similarity, computed as the cosine similarity between the vectors of each protein in a pair. 
Adopting the methodology employed by~\cite{kulmanov2019embeddings} and \cite{xiong2022faithful},
for each positive pair $e_1$ and $e_2$ in the dataset, we compute the similarity between $e_1$ and all other entities and identify the rank of $e_2$. The performance was measured using recall at rank 10\footnote{Since we compute the similarity score for all possible pairs to simulate a more realistic scenario where a user is presented with a ranked list of candidate interactions, the task is several degrees more difficult to performant and all KG embedding methods have a recall score of 0 at rank 1. As a result, we have excluded the results for this metric from our analysis.}, recall at rank 100, mean rank, and the area under the ROC curve (Table \ref{tab:ppiPrediction-sim}). 
Results show that TrueWalksOA achieves the top performance across all metrics, but TrueWalks is bested by RDF2Vec on all metrics except Hits@10, by OWL2Vec* on Hits@100 and by node2vec on Hits@10. 

\begin{table}[!t]
\centering
\footnotesize
\caption{Hits@10, Hits@100, mean rank, and ROC-AUC for PPI prediction using cosine similarity obtained with different methods. In bold, the best value for each metric.}
\begin{tabular}{clcccc}
\toprule
& \textbf{Method} & \textbf{Hits@10}   & \textbf{Hits@100} & \textbf{MeanRank} & \textbf{AUC}   \\ \midrule
\parbox[t]{4mm}{\multirow{9}{*}{\rotatebox[origin=c]{90}{Pos}}} 
& TransE      &   0.013                & 0.125                & 103.934              & 0.538  \\
& TransH     &   0.013                & 0.134                & 102.703              & 0.543\\
& TransR & 0.037	& 0.196 &	81.916	& 0.636 \\
& ComplEx & 0.080 & 0.261 & 64.558 & 0.689 \\
& distMult    &  0.112                & 0.340                & 46.512               & 0.803\\
& DeepWalk & 0.125 & 0.380 & 35.406 & 0.847 \\
& node2vec & 0.163 & 0.375 & 37.275 & 0.827 \\
& metapath2vec & 0.017 & 0.151 & 98.445 & 0.558\\
& OWL2Vec*    &  0.152                & 0.386                & 33.192               & 0.860\\
& RDF2Vec     &   0.133                & 0.391                & 32.419               & 0.870 \\  \midrule
\parbox[t]{4mm}{\multirow{9}{*}{\rotatebox[origin=c]{90}{Pos + Neg}}} 
& TransE       &  0.022                & 0.161                & 94.809               & 0.576 \\
& TransR & 0.100	& 0.274	& 60.120	& 0.732\\
& TransH       &  0.025                & 0.174                & 91.553               & 0.594\\
& ComplEx & 0.132 & 0.334 & 45.268 & 0.805 \\
& distMult   &  0.149                & 0.378                & 35.351               & 0.853\\
& DeepWalk & 0.148 & 0.383 & 35.365 & 0.849 \\
& node2vec & \textbf{0.166} & 0.389 & 34.305 & 0.840\\
& metapath2vec & 0.020 & 0.165 & 93.374 & 0.578\\ 
& OWL2Vec*   &  0.160                & 0.397                & 32.234               & 0.869 \\
& RDF2Vec      &  0.155                & 0.401                & 30.281               & 0.879\\ \midrule
\multicolumn{2}{l}{TrueWalks}   &  0.161                & 0.392                & 32.089               & 0.869  \\
\multicolumn{2}{l}{TrueWalksOA} & \textbf{0.166}       & \textbf{0.407}       & \textbf{28.128}      & \textbf{0.889} \\ 
\bottomrule
\end{tabular}
\label{tab:ppiPrediction-sim}
\end{table}

\begin{figure}[ht!]
    \centering
    \includegraphics[width=\textwidth]{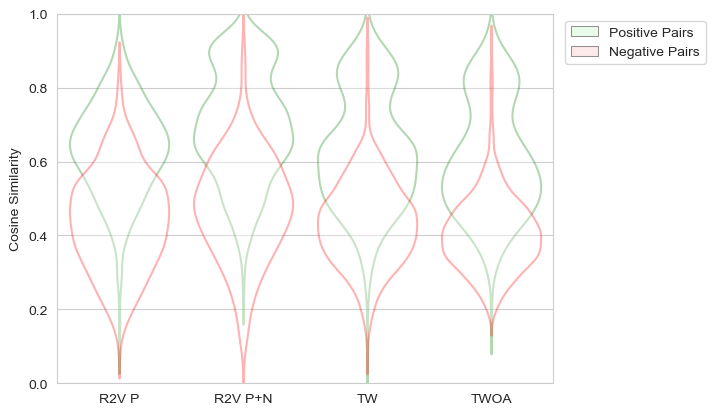}
    \caption{Violin plot with embedding similarity obtained with RDF2Vec with positive statements (R2V P), RDF2Vec with both positive and negative statements (R2V P+N), TrueWalks (TW), and TrueWalksOA (TWOA).}
    \label{fig:ViolinPlot}
\end{figure}

To better understand these results, we plotted the distribution of similarity values for positive and negative pairs in Figure~\ref{fig:ViolinPlot}. There is a smaller overlap between negative and positive pairs similarities for TrueWalksOA, which indicates that considering both the status of the function assignments and the order of entities in the random walks results in embeddings that are more meaningful semantic representations of proteins. 
Furthermore, the cosine similarity for negative pairs is consistently lower when using both variants of TrueWalks, which supports that the contribution of negative statement-based embeddings is working towards filtering out false positives.

\section{Conclusion}

Knowledge graph embeddings are increasingly used in biomedical applications such as the prediction of protein–protein interactions, gene-disease associations, drug-target relations and drug-drug interactions~\cite{mohamed2021biological}. Our novel approach, TrueWalks, was motivated by the fact that existing knowledge graph embedding methods are ill-equipped to handle negative statements, despite their recognized importance in biomedical machine learning tasks~\cite{kulmanov2021semantic}. TrueWalks incorporates a novel walk-generation method that distinguishes between positive and negative statements and considers the semantic implications of negation in ontology-rich knowledge graphs. 
It generates two separate embeddings, one for each type of statement, enabling a dual representation of entities that can be explored by downstream ML, focusing both on features entities have and those they lack.
TrueWalks outperforms representative and state-of-the-art knowledge graph embedding approaches in the prediction of protein-protein interactions and gene-disease associations. 

We expect TrueWalks to be generalizable to other biomedical applications where negative statements play a decisive role, such as predicting disease-related phenotypes~\cite{xue2019predicting} or performing differential diagnosis~\cite{kohler2019encoding}.
In future work, we would also like to explore counter-fitting approaches, such as those proposed for language embeddings~\cite{mrksic2016counter}, to consider how opposite statements can impact the dissimilarity of entities.

\vspace{0.3cm}
\footnotesize
\noindent \textbf{Supplemental Material Statement:} 
The source code for True Walks is available on GitHub (\url{https://github.com/liseda-lab/TrueWalks}). All datasets are available on Zenodo (\url{https://doi.org/10.5281/zenodo.7709195}).
A supplementary file contains the links to the data sources, the parameters for the KG embedding methods and ML models, and the results of the statistical tests.

\vspace{0.3cm}
\footnotesize
\noindent \textbf{Acknowledgements}
C.P., S.S., and R.T.S. are funded by FCT, Portugal, through LASIGE Research Unit (ref. UIDB/00408/2020 and ref. UIDP/00408/2020). R.T.S. acknowledges the FCT PhD grant (ref. SFRH/BD/145377/2019) and DAAD Contact Fellowship grant. This work was also partially supported by the KATY project, which has received funding from the European Union’s Horizon 2020 research and innovation programme under grant agreement No 101017453, and in part by projeto 41, HfPT: Health from Portugal, funded by the Portuguese Plano de Recuperação e Resiliência. The authors are grateful to Lina Aveiro and Carlota Cardoso for the fruitful discussions that inspired this work.

%
\bibliographystyle{splncs04}
\bibliography{references}

\end{document}


\title{Biomedical Knowledge Graph Embeddings with Negative Statements - Supplementary Material}
\author{Rita T. Sousa\inst{1}\orcidID{0000-0002-7241-8970} \and
Sara Silva\inst{1}\orcidID{0000-0001-8223-4799} \and
Heiko Paulheim\inst{2}\orcidID{0000-0003-4386-8195} \and 
Catia Pesquita\inst{1}\orcidID{0000-0002-1847-9393}}
\authorrunning{Sousa et al.}
\institute{LASIGE, Faculdade de Ciências da Universidade de Lisboa, Portugal \\
\email{\{risousa,sgsilva,clpesquita\}@ciencias.ulisboa.pt} \\
\and
Data and Web Science Group, Universität Mannheim, Germany \\ 
\email{heiko.paulheim@uni-mannheim.de}\\
}
\maketitle

\section{Parameters for Knowledge Graph Embedding Methods}

We used nine different knowledge graph embedding approaches as baselines in our work: TransE, TransH, TransR, distMult, DeepWalk, node2vec, metapath2vec, OWL2Vec*, and RDF2Vec. Each method is run with two different KGs, one with only positive statements and one with both positive and negative statements. All of these methods had entity embedding sizes set to 200. 
Table~\ref{tab:GitHubImplementations} contains links to the implementations used for the various methods. For DeepWalk, we modified the RDF2Vec implementation to account for the differences between the two methods. 

\begin{table}[ht!]
    \centering
    \caption{GitHub implementations used for the different methods.}
    \begin{tabular}{ll}
    \toprule
         \textbf{Methods} & \textbf{Implementation} \\ \midrule
         TransE, TransH, TransR, distMult & \url{https://github.com/thunlp/OpenKE} \\
         node2vec & \url{https://github.com/eliorc/node2vec} \\
         metapath2vec & \url{https://github.com/loginaway/Metapath2vec} \\
         OWL2Vec* & \url{https://github.com/KRR-Oxford/OWL2Vec-Star}   \\
        RDF2Vec & \url{https://github.com/IBCNServices/pyRDF2Vec} \\
    \bottomrule
    \end{tabular}
    \label{tab:GitHubImplementations}
\end{table}

The parameters for TransE, TransH, TransR, and distMult are shown in Table~\ref{tab:openke-parameters} and were selected as they are provided as examples in the OpenKE toolkit.

\begin{table}[h!]
    \centering
    \caption{Parameters for TransE, TransH, TransR, and distMult.}
    \begin{tabular}{lccccc}
    \toprule
         \textbf{Parameter} & \textbf{TransE} & \textbf{TransH} & \textbf{TransR} & \textbf{distMult} \\ \midrule
         Train times & 500 & 500 & 500 & 500 \\
         Number batches & 100 & 100 & 100 & 100 \\
         Learning rate & 0.001 & 0.001 & 0.001 & 0.05 \\
         Optimization method & SGD & SGD & SGD & Adagrad \\
         Embedding size & 200 & 200 & 200 & 200 \\
         Entity neg rate & 1 & 1 & 1 & 1 \\
         Relation neg rate & 0 & 0 & 0 & 0 \\
         Bern & 1 & 0 & -- & 0 \\
         Lambda & 0.05 & -- & 4 & --\\
         Margin & -- & 1 & 1 & 1 \\
    \bottomrule
    \end{tabular}
    \label{tab:openke-parameters}
\end{table}

The parameters used for DeepWalk, node2vec, metapath2vec, RDF2Vec, OWL2Vec* are outlined in Table~\ref{tab:walk-based-parameters}. Additional parameters for OWL2Vec* are provided in Table~\ref{tab:owl2vec-parameters}. However, to the best of our knowledge, the option of declaring inverse axioms is not implemented. 

\begin{table}[h!]
    \centering
    \caption{Parameters for methods based on random walks: node2vec, metapath2vec, RDF2Vec, OWL2Vec*, and TrueWalks.}
    \begin{tabular}{lc}
    \toprule
    \textbf{Parameter} & \textbf{Value} \\ \midrule
    Embedding size & 200 \\
    Maximum number of walks per entity & 100 \\
    Maximum depth of walks & 4 \\
    Word2vec model & skip-gram \\
    Epochs for Word2vec model & 5 \\
    Window for Word2vec model & 5 \\
    Minimum count for Word2vec model & 1 \\
    Optimization for softmax & negative sampling\\
    Number of noise words drawn in negative sampling & 5 \\
    Learning rate & 0.025 \\
    \bottomrule
    \end{tabular}
    \label{tab:walk-based-parameters}
\end{table}

\begin{table}[h!]
    \centering
    \caption{Parameters for OWL2Vec*.}
    \begin{tabular}{lc}
    \toprule
         \textbf{Parameter} & \textbf{Value} \\ \midrule
        Ontology projection & No \\
        Project only taxonomy & No \\
        Avoiding OWL constructs & No \\
        Axiom reasoner & None\\
        URI Doc & Yes \\
        Lit Doc & Yes \\
        Mix Doc & No \\
    \bottomrule
    \end{tabular}
    \label{tab:owl2vec-parameters}
\end{table}

\section{Parameters for Machine Learning}

To run the random forest classifier, we employed scikit-learn~\cite{scikit-learn} to optimize certain parameters, specifically the number of estimators and the maximum depth of the tree. The parameters are supplied in Table~\ref{tab:random-forest-parameters}.

\begin{table}[h!]
    \centering
    \caption{Parameters for Random Forest Classifier.}
    \begin{tabular}{lc}
    \toprule
         \textbf{Parameter} & \textbf{Value} \\ \midrule
        Number of estimators & 50,100,200 \\
        Criterion to measure the quality of a split & Gini impurity\\
        Maximum depth of the tree & 2,4,6, None \\
        Number of samples required to split & 2 \\
        Minimum number of samples required to be at a leaf node & 1 \\        
    \bottomrule
    \end{tabular}
    \label{tab:random-forest-parameters}
\end{table}

\section{Data sources}

The versions (and the link where available) of the files used to construct the KGs are listed below:
\begin{itemize}
    \item The GO was downloaded on September 2021 and is available at~\url{http://release.geneontology.org/2021-09-01/ontology/index.html};
    \item The GO positive annotations were downloaded on January 2021 and is available at~\url{http://release.geneontology.org/2021-01-01/annotations/index.html};
    \item The GO negative annotations were downloaded from \url{https://lab.dessimoz.org/20\_not};
    \item The HP was downloaded on October 2022, but a link to this version is no longer available. The version we employed is given as part of the supplementary data and code;
    \item The HP annotations were downloaded on November 2021, but a link to this version is also no longer available. The version we employed is given as part of the supplementary data and code.
\end{itemize}

\section{Statistical Tests}

Statistically significant differences between TrueWalks and the other methods are determined using the non-parametric Wilcoxon test at \mbox{$p < $ 0.05}.
Tables~\ref{tab:p-values5TrueWalks} and \ref{tab:p-values5TrueWalksOA} display the $p$-values obtained from comparing the other methods with TrueWalks and TrueWalksOE, respectively.
TrueWalks performance values are italicized/underlined in Table 2 of the paper when improvements over all other methods are statistically significant, except when comparing TrueWalks with OWL2Vec* for GDA, since in this particular case the improvement is not statistically significant.

\begin{table}
\centering
\caption{$p$-values obtained from comparing TrueWalks with the other methods.}
\begin{tabular}{clccccccc}
\toprule
& \multirow{2}{*}{\textbf{Method}} & \multicolumn{3}{c}{\textbf{PPI Prediction}} && \multicolumn{3}{c}{\textbf{GDA Prediction}}\\
\cmidrule{3-5} \cmidrule{7-9}
& & \textbf{Precision}   & \textbf{Recall} & \textbf{F-measure} & & \textbf{Precision}   & \textbf{Recall} & \textbf{F-measure} \\ \midrule
\parbox[t]{3mm}{\multirow{8}{*}{\rotatebox[origin=c]{90}{Pos}}} 
& TransE      & 0.0000 & 0.0000 & 0.0000 &  & 0.0000 & 0.0000 & 0.0000\\
& TransH     & 0.0000 & 0.0000 & 0.0000 &  & 0.0000 & 0.0002 & 0.0000 \\
& TransR & 0.0000 & 0.0000 & 0.0000 &  & 0.0000 & 0.0734 & 0.0000\\
& distMult   & 0.0000 & 0.0000 & 0.0000 &  & 0.0000 & 0.0145 & 0.0000\\
& DeepWalk & 0.0000 & 0.9997 & 0.0000 &  & 0.0000 & 0.3457 & 0.0004\\
& node2vec & 0.0000 & 0.0000 & 0.0000 &  & 0.0205 & 0.0977 & 0.0204\\
& metapath2vec & 0.0000 & 0.0000 & 0.0000 &  & 0.0000 & 0.0000 & 0.0000\\
& OWL2Vec*   &    0.0000 & 0.0003 & 0.0000 &  & 0.0083 & 0.7352 & 0.0532 \\
& RDF2Vec    &   0.0000 & 0.7077 & 0.0000 &  & 0.0000 & 0.0429 & 0.0000   \\

 \midrule
\parbox[t]{3mm}{\multirow{8}{*}{\rotatebox[origin=c]{90}{Pos+Neg}}} 
& TransE       &         0.0000 & 0.0000 & 0.0000 &  & 0.0000 & 0.0100 & 0.0000  \\
& TransH      &      0.0000 & 0.0000 & 0.0000 &  & 0.0000 & 0.0011 & 0.0000     \\
& TransR  & 0.0000 & 0.0000 & 0.0000 &  & 0.0015 & 0.1608 & 0.0025 \\
& distMult    &    0.0000 & 0.0000 & 0.0000 &  & 0.0000 & 0.0000 & 0.0000    \\
& DeepWalk & 0.0000 & 0.9963 & 0.1268 &  & 0.0000 & 0.0000 & 0.0000\\
& node2vec & 0.0004 & 0.0000 & 0.0000 &  & 0.0001 & 0.3681 & 0.0004 \\
& metapath2vec & 0.0000 & 0.0000 & 0.0000 &  & 0.0004 & 0.0222 & 0.0003 \\
& OWL2Vec*    &     0.0044 & 0.1546 & 0.0044 &  & 0.0783 & 0.0889 & 0.0338       \\
& RDF2Vec     & 0.0000 & 1.0000 & 0.4795 &  & 0.0003 & 0.8611 & 0.0012\\
\bottomrule
\end{tabular}
\label{tab:p-values5TrueWalks}
\end{table}

\begin{table}
\centering
\caption{$p$-values obtained from comparing TrueWalksOA with the other methods.}
\begin{tabular}{clccccccc}
\toprule
& \multirow{2}{*}{\textbf{Method}} & \multicolumn{3}{c}{\textbf{PPI Prediction}} && \multicolumn{3}{c}{\textbf{GDA Prediction}}\\
\cmidrule{3-5} \cmidrule{7-9}
& & \textbf{Precision}   & \textbf{Recall} & \textbf{F-measure} & & \textbf{Precision}   & \textbf{Recall} & \textbf{F-measure} \\ \midrule
\parbox[t]{3mm}{\multirow{8}{*}{\rotatebox[origin=c]{90}{Pos}}} 
& TransE       &      0.0000 & 0.0000 & 0.0000 &  & 0.0000 & 0.0000 & 0.0000     \\
& TransH      &   0.0000 & 0.0000 & 0.0000 &  & 0.0000 & 0.0010 & 0.0000        \\
& TransR      &   0.0000 & 0.0000 & 0.0000 &  &  0.0001	& 0.0662	& 0.0001\\
& distMult    &   0.0000 & 0.0000 & 0.0000 &  & 0.0000 & 0.0415 & 0.0001     \\
& DeepWalk &  0.0000 & 0.7415 & 0.0000 &  & 0.0009 & 0.6676 & 0.0074\\
& node2vec & 0.0000 & 0.0000 & 0.0000 &  & 0.0360 & 0.2502 & 0.0272\\
& metapath2vec & 0.0000 & 0.0000 & 0.0000 &  & 0.0000 & 0.0001 & 0.0000 \\
& OWL2Vec*    &     0.0000 & 0.0000 & 0.0000 &  & 0.0239 & 0.9701 & 0.0954       \\
& RDF2Vec     &  0.0000 & 0.0343 & 0.0000 &  & 0.0004 & 0.2224 & 0.0005\\
 \midrule
\parbox[t]{3mm}{\multirow{8}{*}{\rotatebox[origin=c]{90}{Pos+Neg}}} 
& TransE      &  0.0000 & 0.0000 & 0.0000 &  & 0.0001 & 0.0389 & 0.0000\\
& TransH     & 0.0000 & 0.0000 & 0.0000 &  & 0.0000 & 0.0003 & 0.0000\\
& TransR     &   0.0000 & 0.0000 & 0.0000 &  & 0.0083	& 0.3892 & 0.0135 \\
& distMult   & 0.0000 & 0.0000 & 0.0000 &  & 0.0000 & 0.0000 & 0.0000\\
& DeepWalk & 0.0003 & 0.2720 & 0.0005 &  & 0.0000 & 0.0000 & 0.0000\\
& node2vec & 0.0000 & 0.0000 & 0.0000 &  & 0.0005 & 0.6200 & 0.0038\\
& metapath2vec & 0.0000 & 0.0000 & 0.0000 &  & 0.0005 & 0.0376 & 0.0005\\
& OWL2Vec*   &     0.0008 & 0.0016 & 0.0001 &  & 0.1495 & 0.2189 & 0.1311\\
& RDF2Vec    &  0.0000 & 0.8182 & 0.0066 &  & 0.0018 & 0.9091 & 0.0142    \\
\bottomrule
\end{tabular}
\label{tab:p-values5TrueWalksOA}
\end{table}

\bibliographystyle{splncs04}
\bibliography{references}